\pdfoutput=1

\documentclass[11pt]{article}

\usepackage[]{acl}

\usepackage{times}
\usepackage{latexsym}
\usepackage{graphicx}
\usepackage{subfigure}
\usepackage{hyperref}       
\usepackage{algorithmic}
\usepackage{algorithm}
\usepackage{tabularx}

\usepackage[T1]{fontenc}

\usepackage[utf8]{inputenc}

\usepackage{microtype}

\usepackage{inconsolata}

\usepackage{array}
\usepackage{pifont}
\usepackage{tabularx}
\usepackage{adjustbox}
\usepackage{multirow}
\usepackage{enumitem}
\usepackage{xspace}
\usepackage{tcolorbox}
\usepackage{booktabs,amsfonts,dcolumn}
\usepackage{hyperref}
\usepackage{url}
\usepackage{amsmath,amsthm,amsfonts,amssymb,bm,stmaryrd,bbm}
\usepackage[noorphans,vskip=0.75ex,leftmargin=2ex]{quoting}
\usepackage{footmisc}\interfootnotelinepenalty=10000
\usepackage{colortbl}
\usepackage[noabbrev,capitalize]{cleveref}

\newtcolorbox[list inside=prompt,auto counter,number within=section]{prompt}[1][]{
    colbacktitle=black!60,
    coltitle=white,
    fontupper=\footnotesize,
    boxsep=5pt,
    left=0pt,
    right=0pt,
    top=0pt,
    bottom=0pt,
    boxrule=1pt,
    #1,
}

\newcommand\tf[1]{\textbf{#1}}

\renewcommand{\paragraph}[1]{\vspace{0.2cm}\noindent\textbf{#1}}

\newcommand{\vani}{{\sc{Vanilla}}}
\newcommand{\summ}{{\sc{Summ}}}
\newcommand{\ours}{{\sc{Vtg}}}
\newcommand{\snippet}{{\sc{Snippet}}}

\newcommand{\rerank}{{\sc{Rerank}}}

\newcommand{\posthoc}{{\sc{PostCite}}}

\newcommand{\refinecite}{{\sc{RefineCite}}}
\newcommand{\vericite}{{\sc{VeriCite}}}
\newcommand{\verirefine}{{\sc{VeriRefine}}}

\newcommand{\pvani}{Vanilla}
\newcommand{\psumm}{Summ}
\newcommand{\psnippet}{Snippet}

\newcommand{\prerank}{Rerank}

\newcommand{\pposthoc}{PostCite}
\newcommand{\prefinecite}{RefineCite}
\newcommand{\pvericite}{VeriCite}
\newcommand{\pverirefine}{VeriRefine}

\newcommand{\headercolor}{\rowcolor{gray!15}}

\title{Towards Verifiable Text Generation with Evolving Memory\\ and Self-Reflection}

\author{
Hao Sun\textsuperscript{1,2},
Hengyi Cai\textsuperscript{3},
Bo Wang\textsuperscript{5},
Yingyan Hou\textsuperscript{3}\\
\textbf{Xiaochi Wei\textsuperscript{4},
Shuaiqiang Wang\textsuperscript{4},
Yan Zhang\textsuperscript{1,2},
Dawei Yin\textsuperscript{4}}
\\
\textsuperscript{1}State Key Laboratory of General Artificial Intelligence, Peking University, Beijing, China\\
\textsuperscript{2}School of Intelligence Science and Technology, Peking University\\
\textsuperscript{3}Chinese Academy of Sciences, \textsuperscript{4}Baidu Inc, \textsuperscript{5}Beijing Institute of Technology\\
\tt{sunhao@stu.pku.edu.cn}\\
}

\begin{document}
\maketitle
\begin{abstract}
Despite the remarkable ability of large language models (LLMs) in language comprehension and generation, they often suffer from producing factually incorrect information, also known as hallucination.
A promising solution to this issue is verifiable text generation, which prompts LLMs to generate content with citations for accuracy verification.
However, verifiable text generation is non-trivial due to the focus-shifting phenomenon, the intricate reasoning needed to align the claim with correct citations, and the dilemma between the precision and breadth of retrieved documents.
In this paper, we present \textbf{\textsc{VTG}}, an innovative framework for \textbf{V}erifiable \textbf{T}ext \textbf{G}eneration with evolving memory and self-reflection.
\ours{} introduces evolving long short-term memory to retain both valuable documents and recent documents.
A two-tier verifier equipped with an evidence finder is proposed to rethink and reflect on the relationship between the claim and citations. 
Furthermore, active retrieval and diverse query generation are utilized to enhance both the precision and breadth of the retrieved documents.
We conduct extensive experiments on five datasets across three knowledge-intensive tasks and the results reveal that \ours{} significantly outperforms baselines.
\end{abstract}

\section{Introduction}
Large Language Models~(LLMs)~\cite{Scao-arxiv-2022-BLOOM,Taylor-arxiv-2022-Galactica,chowdhery2022palm} have showcased remarkable performance across a spectrum of downstream tasks recently. 
Despite their advancements, LLMs often generate responses that include hallucinated facts and inaccurate information~\cite{ji2023survey,shuster2021retrieval, zhang2023sac3}, undermining their reliability. 

\begin{figure}[t]
    \includegraphics[width=\linewidth]{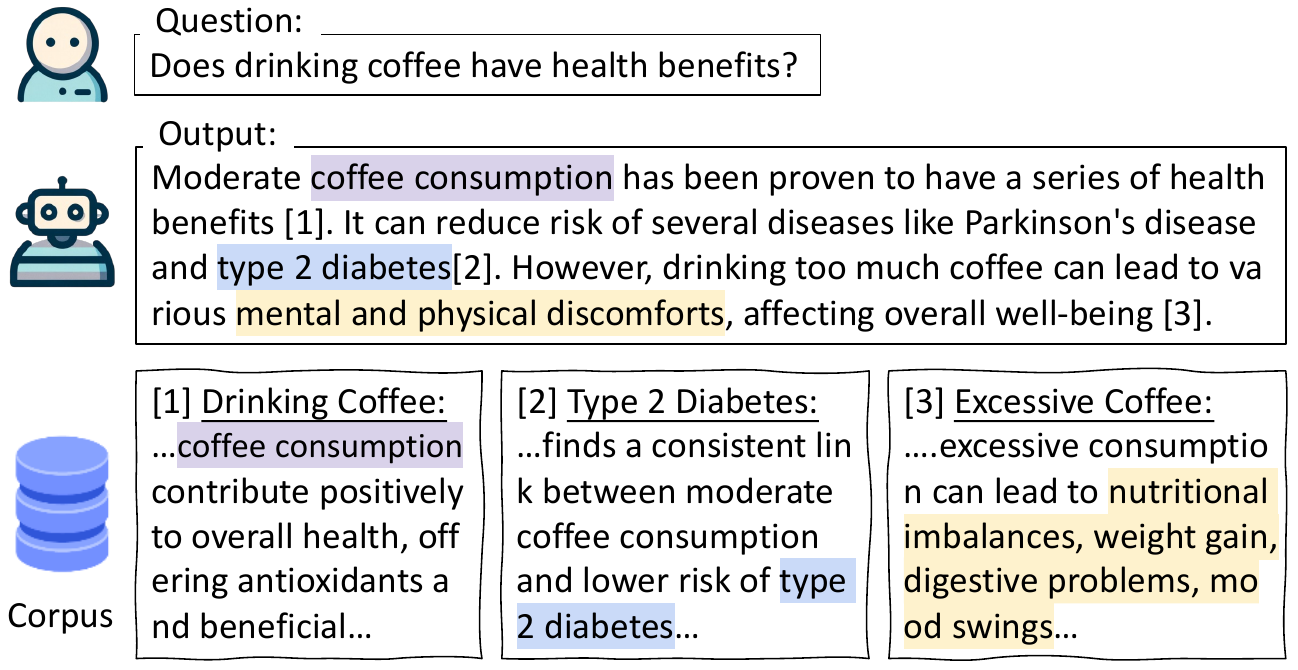}
    \caption{Given a question, the system generates text while providing citing documents from a large corpus.}
    \label{fig:main}
\end{figure}

To enhance the reliability of LLMs, a new generation paradigm, Verifiable Text Generation~\cite{gao2023enabling, gao2022rarr, bohnet2022attributed, liu2023evaluating, li2023survey, funkquist2022citebench}, is proposed to encourage LLMs to provide citations for any claim they generate.
For example, as shown in \cref{fig:main}, the response to the question “Does drinking coffee have health benefits?” contains authentic sources supporting the claims, enhancing its credibility.
In this way, verifiable generation produces more trustworthy answers, which facilitates its application in multiple commercial systems, such as Bing Chat\footnote{\url{https://www.bing.com/new}} and perplexity.ai\footnote{\url{https://www.perplexity.ai}}.

However, verifiable text generation is challenging for the following reasons. 
\textbf{Firstly}, it often involves long text generation, where the focus of the content changes over time, characterized by focus-shifting phenomenon~\cite{lan2021modeling, sun2023history}.
This dynamic poses challenges in consistently aligning claims with the appropriate evidential references.
For instance, as depicted in \cref{fig:main}, the discussion evolves from the health benefits of moderate coffee consumption to the adverse effects of excessive consumption.
Such shifts demand a dynamic adaptation in the document pool to support the shifting focus of the generation.
\textbf{Secondly}, identifying the intricate relationship between a claim and its potential evidence requires more than just linguistic matching—it demands careful analysis.
For example, in \cref{fig:main}, the third claim suggests excessive coffee consumption leads to various mental and physical discomforts.
The corresponding supporting document, although not explicitly mentioning these discomforts, describes symptoms inherently related to them. 
This necessitates an in-depth examination to confirm that the evidence truly supports the specific claim.
\textbf{Thirdly}, striking a balance between the precision and breadth of retrieved documents presents a complex challenge for verifiable text generation.
On the one hand, the task is susceptible to noisy documents during the claim-citation alignment process, emphasizing the need to selectively retain a few highly relevant documents.
On the other hand, the intrinsic nature of verifiable text generation calls for a comprehensive collection of documents to enhance credibility.
Therefore, crafting strategies to balance precision and breadth in document retrieval is crucial for advancing verifiable text generation.

When composing text with citations, individuals are capable of adaptively gathering the most relevant information regarding the claim being written, typically involving active information seeking and frequent verification.
Inspired by this process, we propose \textbf{\textsc{Vtg}}, short for \textbf{V}erifiable \textbf{T}ext \textbf{G}eneration, a novel framework that operates through iterative generation and verification, utilizing an evolving memory and a two-tier verifier.
\text{Specifically}, to address the challenge of focus-shifting, \ours{} employs an evolving long short-term memory system. This system effectively archives important documents in long-term memory and maintains recent ones in short-term memory, thereby providing support for the evolving focus of the generation.
\text{Moreover}, to identify the complex relationship between a claim and its potential evidence, \ours{} employs a generation verifier and a memory verifier, both using Natural Language Inference (NLI) model to assess the logical support of potential evidence for the claim.
The generation verifier first checks if the cited documents logically support the claim. 
If there's a misalignment, the memory verifier reevaluates the claim against the documents stored in memory.
A positive outcome suggests that the misalignment is due to the citation generation process, not because the information in the claim is wrong, leading to the adoption of a refined set of documents from memory for citation.
Conversely, a negative outcome indicates potential factual inaccuracies in the claim, triggering an evidence finder to gather external information, which facilitates the regeneration of a more accurate and verifiable claim.
\text{Lastly}, to balance between precision and breadth in document retrieval, \ours{} incorporates active retrieval and diverse query generation.
Retrieval is initiated only when the claim does not pass the memory verifier, indicating potential factual inaccuracies.
This approach guarantees the necessity of retrieval, reducing noise from unnecessary retrieval, and thereby enhancing retrieval precision. 
By instructing LLMs to generate diverse queries, the breadth of retrieved documents is broadened, enabling the documents to offer comprehensive support for the claim.

To summarize, our main contributions are:
\begin{itemize}
\item We introduce \ours{}, a novel framework that guides the generation model using the combination of an evolving memory and a two-tier verifier, offering an adaptive and reflective approach for verifiable text generation.
\item The evolving memory stores valuable and recent documents, effectively addressing the focus-shifting challenge.
The two-tier verifier and evidence finder enable the in-depth examination of the claim and its potential evidence. 
The active retrieval and diverse query generation can improve both the precision and breadth of the retrieved documents.
\item We conduct extensive experiments on five datasets across three knowledge-intensive tasks and the results show that \ours{} significantly outperforms baselines on both citation quality and answer correctness.
\end{itemize}
\section{Methodology}
In this section, we first present the overall framework of \ours{}.
Then we will go over each part of the model in detail.

\begin{figure*}[tb]
\includegraphics[width=0.85\textwidth]{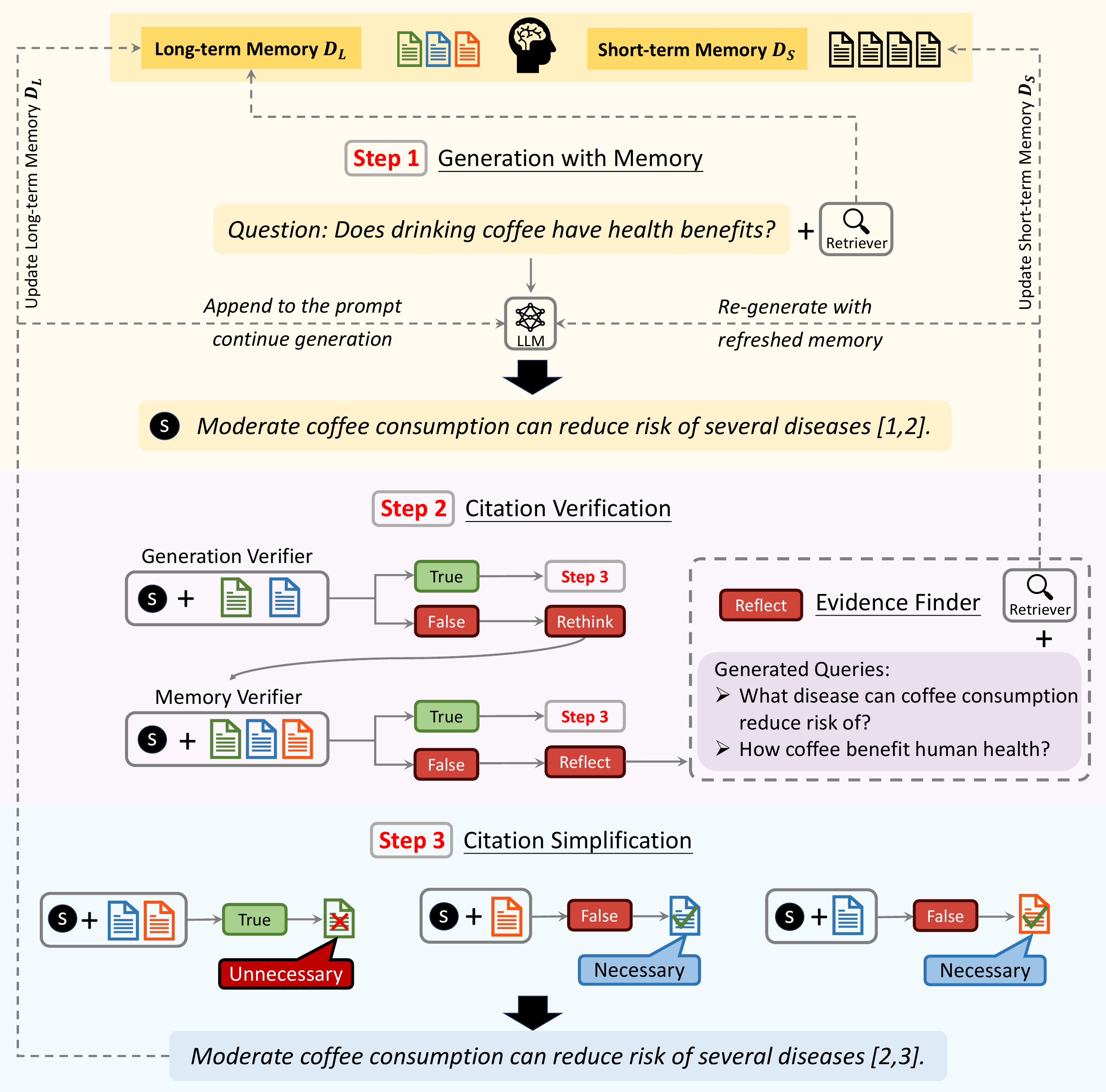}
\centering
\caption{The illustration of \ours{}, which mainly consists of three stages: Generation with Memory, Citation Verification, and Citation Simplification.
The Evidence Finder will only be activated when the claim fails to pass the memory verifier, indicating potential factual inaccuracies in the claim.
}
\label{fig:model}
\end{figure*}

\subsection{Overall Framework}
Given a question $q$ and a corpus of text passages $\mathcal{D}$, the task of verifiable text generation demands the system to return an output $\mathcal{S}$, which consists of $n$ claims, and each claim $s_i$ cites a list of passages $\mathcal{C}_i = \{c_{i,1}, c_{i, 2}, \ldots\}$.
As shown in \cref{fig:model}, \ours{} operates with an evolving memory system: the long-term memory $D_L$ that is maintained throughout the generation process, and the short-term memory $D_S$ that is continually updated to align with the shifting focus of content.
Initially, $D_L$ is filled with the top-$k$ retrieved documents based on the original question, while $D_S$ starts empty. 
The LLM generates the first claim and its corresponding citation based on documents from both memories.
Subsequently, the generation verifier examines whether the cited documents logically support the claim.
If the outcome is negative, the memory verifier then evaluates whether the combined memories (i.e., $D_S \cup D_L$) logically support the claim.
If either the cited documents or the documents in the memories support the claim, the corresponding document set will undergo simplification and be incorporated into $D_L$ for future generations.
Conversely, if neither of them supports the claim, the LLM generates diverse queries about the claim, which are then used to retrieve documents to refresh $D_S$.
In cases where a regenerated claim fails to pass verification after $T$ attempts, the model progresses to the next stage of generation.
This interactive and iterative process continues until the LLM generates an end token, indicating generation completion.

\vspace{-1em}
\subsection{Generation with Evolving Memory}
In verifiable text generation, the content’s focus may shift as the generation progresses, known as the focus-shifting phenomenon~\cite{lan2021modeling, sun2023history}.
Consequently, the document pool must be dynamically adjusted to ensure that the documents can support the latest claim.
This variability presents a challenge when generating claims and citations simultaneously due to fluctuating reference indices in previously generated content, which could potentially bring confusion into the model’s generation process.
Moreover, insights from previous research~\cite{juneja2023small} suggest that tasks requiring distinct skills are more effectively handled through separate modules instead of one monolithic module.
Based on these insights, \ours{} adopts a divided approach: dividing the verifiable text generation task into distinct claim generation and citation generation processes.

\paragraph{Claim Generator} aims to complete unfinished content based on the documents from both long-term memory $D_L$ and short-term memory $D_S$. 
Initially, long-term memory $D_L$ is filled with top-$k$ documents selected based on their relevance to the original question.
This ensures that information closely related to the original question is permanently retained in memory, providing a consistent guide throughout the entire generation process.
In contrast, short-term memory $D_S$ starts as an empty set, ready to be dynamically filled as needed.

\paragraph{Citation Generator} aims to source supporting evidence for the generated claim from both long-term memory $D_L$ and short-term memory $D_S$. 
This focused approach allows the citation generator to concentrate on validating the current claim without being distracted by unrelated contexts.

\subsection{Citation Construction}
Model hallucination in LLM outputs often leads to inaccuracies and distortions~\cite{ji2023survey,shuster2021retrieval, zhang2023sac3}. One approach to mitigate this issue is by grounding the model's outputs in specific documents. However, even citations generated by LLM can be subject to hallucinations, potentially leading to unreliable or irrelevant references.
To ensure that the evidence truly supports the specific claim, \ours{} employs a two-tier verification system comprising the generation verifier and the memory verifier, both of which use Natural Language Inference (NLI) model to assess the logical support of potential evidence (premise) for the claim (hypothesis).
The generation verifier assesses whether the cited documents logically support the claim, while the memory verifier evaluates the support provided by the entire memory set.
Upon successful verification by either component, the citation set advances to the citation simplifier to remove any unnecessary references.

\paragraph{Generation Verifier} examines whether the citation set logically supports the claim.
If the verification is successful, the citation set advances to the citation simplifier for further refinement. 
However, if the verification fails, there are two possible reasons:
1) The claim is logically consistent with the memory set, but the generated citation set is inaccurate. In this case, the citation simplifier removes unnecessary documents from the full memory set, and the remaining documents are used as the citation.
2) The claim lacks support from the memory set, suggesting potential factual inaccuracies in the claim. Here, the evidence finder is activated to seek relevant information for claim regeneration.

\paragraph{Memory Verifier} detects the potential hallucination of the current claim by analyzing whether the full memory set logically supports it.
A positive outcome suggests that the claim-citation misalignment originates from the citation generation process, triggering the citation simplifier to refine the full memory set to be used as the final citation. 
However, if the full memory set still cannot validate the current claim, it implies that the LLM may have fabricated the claim with its parametric knowledge and the claim might be factually incorrect, necessitating the employment of the evidence finder to seek evidence for claim regeneration.
By doing so, the proposed framework is able to assess the generated citations in a self-reflection manner.

\paragraph{Citation Simplifier} is designed to eliminate unnecessary references. 
It works by iteratively reviewing each document in the citation set, temporarily removing one, and assessing if the claim remains well-supported without it.
Redundant citations, that do not contribute to the claim's verification, are permanently removed. 
This iterative process ensures the final citation set is concise and supportive, retaining only essential citations for claim verification.
These citations, after verification and simplification, become part of long-term memory $D_L$ to guide future claim generations.

\subsection{Evidence Finder}
When the full memory set fails to pass the memory verifier, indicating potential factual inaccuracies in the generated claim, the evidence finder is activated to retrieve relevant documents for verification.

The intrinsic nature of verifiable text generation necessitates a wide range of documents to boost the credibility of the generated content.
To expand the knowledge scope of the retrieved documents, \ours{} prompt LLMs to formulate queries that explore various aspects of the current claim.
Furthermore, traditional query generation methods, which typically rely solely on the current claim, can lead to ambiguity, particularly with claims containing pronouns or unclear references.
To overcome this, \ours{} introduces a context-aware query generation approach. This method enhances query generation by incorporating the original question, the current claim and the unfinished content into the prompt. 

Once queries are generated, a retriever collects relevant documents for each query.
These documents then update the short-term memory $D_S$, with the latest and most relevant information.
This key update ensures that the short-term memory is both current and pertinent, thereby enabling LLM to provide precise citations for the latest claim.
\section{Experiment}

\begin{table*}[tb]
\centering
\resizebox{\linewidth}{!}{
\begin{tabular}{lcccccccccccccccccccccc}
\toprule
\textbf{Datasets} & \multicolumn{8}{c}{\textbf{Wikihop}} & 
\multicolumn{7}{c}{\textbf{WebQ}}  & 
\multicolumn{7}{c}{\textbf{NQ}}  \\
 \cmidrule(lr){2-9}  \cmidrule(lr){10-16}  \cmidrule(lr){17-23} 
& \multicolumn{2}{c}{\tf{Correct}}  & \multicolumn{3}{c}{\tf{ALCE.Citation} } & \multicolumn{3}{c}{\tf{LLM.Citation} } &\tf{Correct} & \multicolumn{3}{c}{\tf{ALCE.Citation} } & \multicolumn{3}{c}{\tf{LLM.Citation} } & \tf{Correct} & \multicolumn{3}{c}{\tf{ALCE.Citation}} & \multicolumn{3}{c}{\tf{LLM.Citation} } \\

\cmidrule(lr){2-3} \cmidrule(lr){4-6}  \cmidrule(lr){7-9}  \cmidrule(lr){10-10}  \cmidrule(lr){11-13}   \cmidrule(lr){14-16} \cmidrule(lr){17-17} \cmidrule(lr){18-20}   \cmidrule(lr){21-23} 

\textbf{Metrics} & \textbf{EM} & \textbf{F1} & \textbf{Rec} & \textbf{Prec} & \textbf{F1} & \textbf{Rec} & \textbf{Prec} & \textbf{F1} & \textbf{EM} & \textbf{Rec} & \textbf{Prec} & \textbf{F1} & \textbf{Rec} & \textbf{Prec} & \textbf{F1} & \textbf{EM} & \textbf{Rec} & \textbf{Prec} & \textbf{F1} & \textbf{Rec} & \textbf{Prec} & \textbf{F1} \\
\midrule
\headercolor
\multicolumn{23}{c}{\textbf{Vicuna-13B}} \\
\textsc{\pvani{}} &   23.40 & 21.98 & 29.55 & 22.25 & 25.39 & 41.59 & 35.04 & 38.03 & 55.80 & 67.66 & 60.66 & 63.97 & 67.50 & 67.83 & 67.67 & 54.80 & 71.39 & 61.71 & 66.20 & 77.46 & 63.94 & 70.05  \\
\summ{} &  23.20 & 20.00 & 30.89 & 28.43 & 29.61 & 37.66 & 39.12 & 38.37 & 58.00 & 70.51 & 62.07 & 66.02 & 68.23 & 66.57 & 67.39 & 57.00 & 51.55 & 52.21 & 51.88 & 56.76 & 62.86 & 59.66   \\
\snippet{} & 21.80 & 20.05 & 25.18 & 21.95 & 23.45 & 33.67 & 29.75 & 31.59 & 58.40 & 53.44 & 49.15 & 51.21 & 68.46 & 69.57 & 69.01 & 57.20 & 43.56 & 41.43 & 42.47 & 57.56 & 59.57 & 58.55  \\
\rerank{} &  22.60 & 21.13 & 47.03 & 47.53 & 47.28 & 54.33 & 53.73 & 54.03 & 56.40 & 89.93 & 76.33 & 82.57 & 88.20 & 67.77 & 76.64 & 56.20 & 83.56 & 73.57 & 78.25 & 81.66 & 73.49 & 77.36   \\
\ours{} &\textbf{ 25.60} & \textbf{23.27} & \textbf{55.36} & \textbf{49.59} & \textbf{52.32} & \textbf{62.76} &\textbf{ 54.69} & \textbf{58.45} & \textbf{60.00} & \textbf{92.16} &\textbf{ 86.51} & \textbf{89.25} & \textbf{89.43} & \textbf{81.38} & \textbf{85.21} &\textbf{ 58.00 }& \textbf{88.69} & \textbf{82.02} &\textbf{ 85.22} & \textbf{86.35} & \textbf{78.06 }& \textbf{82.00}   \\
\midrule
\headercolor
\multicolumn{23}{c}{\textbf{Text-Davinci-003}} \\
\vani{} &  33.00 & 33.01 & 40.46 & 28.30 & 33.30 & 59.07 & 43.00 & 49.77 & 67.50 & 63.78 & 58.97 & 61.28 & 71.18 & 66.52 & 68.77 & 62.50 & 60.48 & 55.56 & 57.92 & 66.45 & 61.59 & 63.93   \\
\summ{} &  30.00 & 30.63 & 9.39 & 12.19 & 10.61 & 23.19 & 24.64 & 23.89 & 67.50 & 60.06 & 47.62 & 53.12 & 68.33 & 56.65 & 61.95 & 62.50 & 44.23 & 38.45 & 41.14 & 55.36 & 49.20 & 52.10    \\
\snippet{} &  32.00 & 30.13 & 13.86 & 18.49 & 15.84 & 37.36 & 38.99 & 38.16 & 67.00 & 65.41 & 52.32 & 58.14 & 71.81 & 68.07 & 69.89 & 62.00 & 54.72 & 46.99 & 50.56 & 73.05 & 69.55 & 71.25   \\
\rerank{} &  32.67 & 33.09 & 56.13 & 45.22 & 50.09 & 63.43 & 46.32 & 53.54 & 67.00 & 73.72 & 64.90 & 69.03 & 78.12 & 70.00 & 73.84 & 61.50 & 71.30 & 63.44 & 67.14 & 79.03 & 66.37 & 72.15  \\
\ours{} &  \textbf{41.50} & \textbf{40.19} & \textbf{63.89} & \textbf{57.65} & \textbf{60.61} & \textbf{70.47} & \textbf{59.13} & \textbf{64.30} & \textbf{68.00} & \textbf{93.00} & \textbf{88.72} & \textbf{90.81} & \textbf{90.70} & \textbf{87.52} & \textbf{89.08} & \textbf{63.00} & \textbf{91.85} & \textbf{86.59} & \textbf{89.14} & \textbf{84.92} & \textbf{73.20} & \textbf{78.63}   \\
\bottomrule
\end{tabular}
}
\caption{Comparisons between \ours{} and baselines on Multi-hop QA task and Open-domain QA task.}
\label{tab:main_result_1}
\end{table*}
\begin{table*}[tb]
\centering
\resizebox{\linewidth}{!}{
\begin{tabular}{lcccccccccccccccccccc}
\toprule
\textbf{Datasets} & \multicolumn{10}{c}{\textbf{ASQA}} & 
\multicolumn{7}{c}{\textbf{ELI5}} & 
\multicolumn{2}{c}{\textbf{Overall}}  \\
\cmidrule(lr){2-11} \cmidrule(lr){12-18} \cmidrule(lr){19-20}
& \multicolumn{4}{c}{\tf{Correct}}  & \multicolumn{3}{c}{\tf{ALCE.Citation} } & \multicolumn{3}{c}{\tf{LLM.Citation} } &\tf{Correct} & \multicolumn{3}{c}{\tf{ALCE.Citation}} & \multicolumn{3}{c}{\tf{LLM.Citation} } & \tf{Correct} & \tf{Citation}  \\

\cmidrule(lr){2-5} \cmidrule(lr){6-8} \cmidrule(lr){9-11} \cmidrule(lr){12-12} \cmidrule(lr){13-15} \cmidrule(lr){16-18}  \cmidrule(lr){19-19} \cmidrule(lr){20-20}
\textbf{Metrics} & \textbf{EM} & \textbf{D-F1} & \textbf{R-L} & \textbf{DR} & \textbf{Rec} & \textbf{Prec} & \textbf{F1} & \textbf{Rec} & \textbf{Prec} & \textbf{F1} & \textbf{Claim} & \textbf{Rec} & \textbf{Prec} & \textbf{F1}  &  \textbf{Rec} & \textbf{Prec} & \textbf{F1} &\textbf{EM}  & \textbf{F1} \\
\midrule
\headercolor
\multicolumn{20}{c}{\textbf{Vicuna-13B}} \\
\vani{} &  32.00 & 27.52 & 33.53 & 30.53 & 72.78 & 62.09 & 67.01 & 73.28 & 66.59 & 69.78 & 12.20 & 59.79 & 48.26 & 53.41 & 81.46 & 76.89 & 79.11 & 35.64 & 60.06   \\
\summ{}  &  41.71 & 28.95 & 37.18 & 33.07 & 62.15 & 59.60 & 60.85 & 68.95 & 70.19 & 69.56 & 14.20 & 60.13 & 52.42 & 56.01 & 77.87 & 72.42 & 75.05 & 38.82 & 57.44   \\
\snippet{} &  39.22 & 27.01 & 35.65 & 31.33 & 46.23 & 47.04 & 46.63 & 56.55 & 63.03 & 59.61 & 14.33 & 31.47 & 32.72 & 32.08 & 46.69 & 49.32 & 47.97 & 38.19 & 46.26    \\
\rerank{} &  37.14 & 28.21 & 32.18 & 30.20 & 88.29 & 75.74 & 81.53 & 88.29 & 75.74 & 81.53 & 11.67 & 73.80 & 61.12 & 66.86 & 84.57 & 77.09 & 80.65 & 36.80 & 72.67   \\
\ours{} & \textbf{41.92} & \textbf{30.53} & \textbf{37.87} & \textbf{34.20} & \textbf{89.15} & \textbf{82.57} & \textbf{85.73} & \textbf{89.15} & \textbf{82.57} & \textbf{85.73} & \textbf{14.73} & \textbf{81.50} & \textbf{72.16} & \textbf{76.55} & \textbf{87.60} & \textbf{84.46} & \textbf{86.00}& \textbf{40.05} & \textbf{78.65}   \\
\midrule
\headercolor
\multicolumn{20}{c}{\textbf{Text-Davinci-003}} \\
\vani{} &  40.25 & 31.47 & 35.81 & 33.64 & 58.13 & 55.17 & 56.61 & 58.13 & 55.17 & 56.61 & 13.43 & 58.66 & 47.40 & 52.43 & 58.66 & 47.40 & 52.43 & 43.34 & 55.31   \\
\summ{} &  41.33 & 28.91 & 37.21 & 33.06 & 48.31 & 40.68 & 44.17 & 50.48 & 44.44 & 47.27 & 11.50 & 39.43 & 31.81 & 35.21 & 52.27 & 48.47 & 50.30 & 42.57 & 41.98  \\
\snippet{} &  39.60 & 30.11 & 38.35 & 34.23 & 53.14 & 43.19 & 47.65 & 59.31 & 52.05 & 55.44 & 13.67 & 45.29 & 37.23 & 40.87 & 62.39 & 55.19 & 58.57 & 42.85 & 50.64   \\
\rerank{} &  39.55 & 29.94 & 39.38 & 34.66 & 75.83 & 69.81 & 72.70 & 76.41 & 70.01 & 73.07 & 14.76 & 76.21 & 61.67 & 68.17 & 86.98 & 77.64 & 82.04 & 43.10 & 68.18   \\
\ours{} & \textbf{41.53} & \textbf{31.64} & \textbf{39.45} & \textbf{35.55} & \textbf{86.70} & \textbf{79.95} & \textbf{83.19} & \textbf{89.10} & \textbf{79.84} & \textbf{84.22} & \textbf{16.67} & \textbf{82.63} & \textbf{71.56} & \textbf{76.70} & \textbf{87.94} & \textbf{81.79} & \textbf{84.75} & \textbf{46.14} & \textbf{80.14}   \\
\bottomrule
\end{tabular}
}
\caption{Comparisons between \ours{} and baselines on Long-form QA task and overall performance.}
\label{tab:main_result_2}
\end{table*}

\subsection{Baselines}
For an equitable comparison, we have selected the following four best-performing baseline methodologies as proposed in ALCE \cite{gao2023enabling}.

\paragraph{\textsc{\pvani{}}:}
The LLM generates responses with citations based on the top-ranked documents.

\paragraph{\textsc{\psumm{}}:}
The LLM first summarizes information from the top-ranked documents and then generates texts with citations based on the summarization.

\paragraph{\textsc{\psnippet{}}:}
The LLM first extracts relevant snippets from the top-ranked documents and then generates texts with citations based on the snippets.

\paragraph{\textsc{\prerank{}}:}
The LLM first generates four unique responses using high temperature and outputs the one with the highest citation recall.

Besides, we also compare with several post-processing baselines, which include \textbf{\textsc{\pposthoc{}}}, \textbf{\textsc{\prefinecite{}}}, \textbf{\textsc{\pvericite{}}} and \textbf{\textsc{\pverirefine{}}}.
Due to space limits, we only put the four baselines from ALCE in the main results, for the complete experiment results, please refer to \cref{baselines}.

\subsection{Datasets and Evaluation}
We assess the effectiveness of our methods on five datasets across three knowledge-intensive tasks. 
For all datasets, our evaluation criteria encompass both the answer correctness and citation quality of model outputs.
The details of the tasks and the datasets we used are as follows:

\paragraph{Multihop QA} entails answering complex questions that necessitate multiple retrieval and reasoning steps \cite{hotpotqa-yang-2018,2wikimultihopqa-ho-2020}. We employ the 2WikiMultihopQA dataset \cite{2wikimultihopqa-ho-2020}, which consists of 2-hop complex questions derived from Wikipedia, requiring skills in composition, comparison or inference.

In line with \citet{jiang2023active}, LLMs are prompted to provide the final answer, which is then evaluated against the reference answer using answer-level Exact Match (EM), token-level precision, recall and F1 metrics.

\paragraph{Long-form QA} aims to generate detailed answers to complex questions \cite{eli5-fan-2019,fan-etal-2019-eli5}, we choose the ASQA dataset \cite{asqa-stelmakh-2022} and the ELI5 dataset \cite{fan-etal-2019-eli5} for evaluation. ASQA focuses on ambiguous questions requiring comprehensive answers covering multiple interpretations. ELI5, on the other hand, deals with complex questions demanding lengthy, in-depth answers backed by multiple documents.

For ASQA, we apply the metrics outlined in \citet{asqa-stelmakh-2022}, including Exact Match (EM), a soft match using a RoBERTa-based QA model (Disambig-F1), ROUGE \cite{lin-2004-rouge} and a combined DR score. In the case of ELI5, we adhere to the evaluation criteria of \citet{gao2023enabling}, focusing on whether the model's predictions address the sub-claims of the gold-standard answer.

\paragraph{Open-domain QA} requires leveraging external knowledge for answering questions. We choose the NQ dataset \cite{kwiatkowski2019natural} and the WebQ dataset \cite{berant2013semantic} for evaluation.

Following the methodology of \citet{yu2022generate} and \citet{sun2023allies}, LLMs are prompted to generate the final answer, which is then compared with the reference answer using answer-level Exact Match (EM).

\paragraph{Verifiability Evaluation.} 
To evaluate the citation quality of responses, we employ the approach of \citet{gao2023enabling}, focusing on calculating \textit{ALCE.Citation Recall}, \textit{ALCE.Citation Precision}, and the combined \textit{ALCE.Citation F1} score. \textit{ALCE.Citation Recall} examines whether the output is fully supported by the cited documents, while \textit{ALCE.Citation Precision} assesses the redundancy of the citations included.
Additionally, to further improve the robustness of the evaluation, we incorporate the use of LLM as the citation evaluator. Specifically, The Qwen-Max is given a sentence and all the passages that the sentence cited and is asked to judge whether the passages fully support the sentence, which is used as the signal to compute \textit{LLM.Citation Recall} and \textit{LLM.Citation Precision}.

For more details on dataset statistics and evaluation details, please refer to \cref{ap:setting}. 

\subsection{Implementation Details}
To prove the generalizability of our method, we conduct experiments using LLMs of different parameter sizes. 
Specifically, we utilize two LLMs: \texttt{Vicuna-13B-v1.5-16k}\footnote{\scriptsize{\url{https://huggingface.co/lmsys/vicuna-13b-v1.5-16k}}}~\cite{zheng2023judging} and \texttt{Text-Davinci-003}\footnote{\scriptsize{\url{https://api.openai.com/v1/completions} as of October 2023}}~\cite{ouyang2022training} for evaluation, respectively.
For the evaluation of verifiability and the inference tasks in both the \rerank{} and \ours{} methods, we employ the TRUE\footnote{\scriptsize{\url{https://huggingface.co/google/t5_xxl_true_nli_mixture}}} model \citep{raffel2020exploring}, a T5-11B model fine-tuned on a collection of NLI datasets, to automatically examine whether the cited documents entail the claim.
We also experiment with different NLI models for inference and evaluation, please refer to \cref{sec:nli} for more details.
Following \citet{gao2023enabling}, we use Wikipedia dump from Dec. 20, 2018 as our retrieval corpus and use DPR~\cite{karpukhin-etal-2020-dense} as our dense retriever.

\subsection{Main Results}
In this section, we present a comparison of the performance of \ours{} against other baselines across five different datasets in \cref{tab:main_result_1,tab:main_result_2}. 
Based on these results, several observations can be made:

First, our proposed \ours{} consistently outperforms other approaches across various datasets and metrics when applied to LLMs of different parameter sizes. 
Notably, \ours{} achieves a significant enhancement in citation quality, with a notable 22\% and 9\% relative improvement over the strongest competitor \rerank{}, when evaluated with \texttt{Text-Davinci-003} and \texttt{Vicuna-13B}, respectively.
Moreover, \ours{}'s strong capability for verifiable generation also leads to a considerable improvement in answer correctness, evidenced by an approximate 5\% overall improvement compared to the leading baselines across both LLMs.

Second, among the evaluated baselines, the method \rerank{} stands out in the aspect of citation quality, primarily owing to its multiple sampling strategy that enhances the chances of producing high-quality outputs. 
However, its performance in answer correctness fluctuates across different datasets, which is notably lower in the ASQA dataset when evaluated with \texttt{Text-Davinci-003} and in the ELI5 dataset when evaluated with \texttt{Vicuna-13B}.
In contrast, \ours{} demonstrates consistent improvement across all metrics and datasets, highlighting its robustness and reliability.

Third, the comparison between the two different LLMs reveals interesting findings. 
By integrating a broader range of documents, \summ{} and \snippet{} outperform \vani{} in terms of overall correctness when evaluated with \texttt{Vicuna-13B}.
However, this advantage diminishes when evaluated with \texttt{Text-Davinci-003}.
This could be attributed to \texttt{Text-Davinci-003}'s extensive internal knowledge base, which enables it to generate answers without relying on external sources, as evidenced by the performance comparison when applying the same method to both LLMs. 
Consequently, \summ{} and \snippet{} may introduce unnecessary noisy information in the context of \texttt{Text-Davinci-003}, leading to correctness degradation.
Additionally, these methods struggle with citation quality on both LLMs, as simplified documents make it hard for LLMs to generate the correct citations.

\begin{table}[t]
\centering
\resizebox{\linewidth}{!}{
\begin{tabular}{lccccc}
\toprule
 & \multicolumn{2}{c}{\textbf{Correct}} & 
\multicolumn{3}{c}{\textbf{Citation}}   \\
\cmidrule(lr){2-3} \cmidrule(lr){4-6} 
& \tf{EM} & \tf{F1} & \tf{Rec} & \tf{Pre} & \tf{F1} \\
\midrule
\ours{} & \textbf{25.60} & \textbf{23.27} &  \textbf{55.36} & \textbf{49.59} & \textbf{52.32} \\
-w/o Verifier & 20.60&17.75& 37.34&30.16&33.36 \\
-w/o Memory & 21.40&18.88& 43.88&35.18&39.05 \\
-w/o Simplifier & 24.20&22.85&45.07&28.61&36.00 \\
-w/o Diverse QG & 21.40&18.99& 47.76&40.15&43.62 \\
\bottomrule
\end{tabular}
}
\caption{Ablation Study on 2WikiMultihopQA.}
\label{tab:ablation}
\end{table}

\begin{figure}[t]
    \centering
    \subfigure[Quey Generation Num]{
    \includegraphics[width=0.22\textwidth]{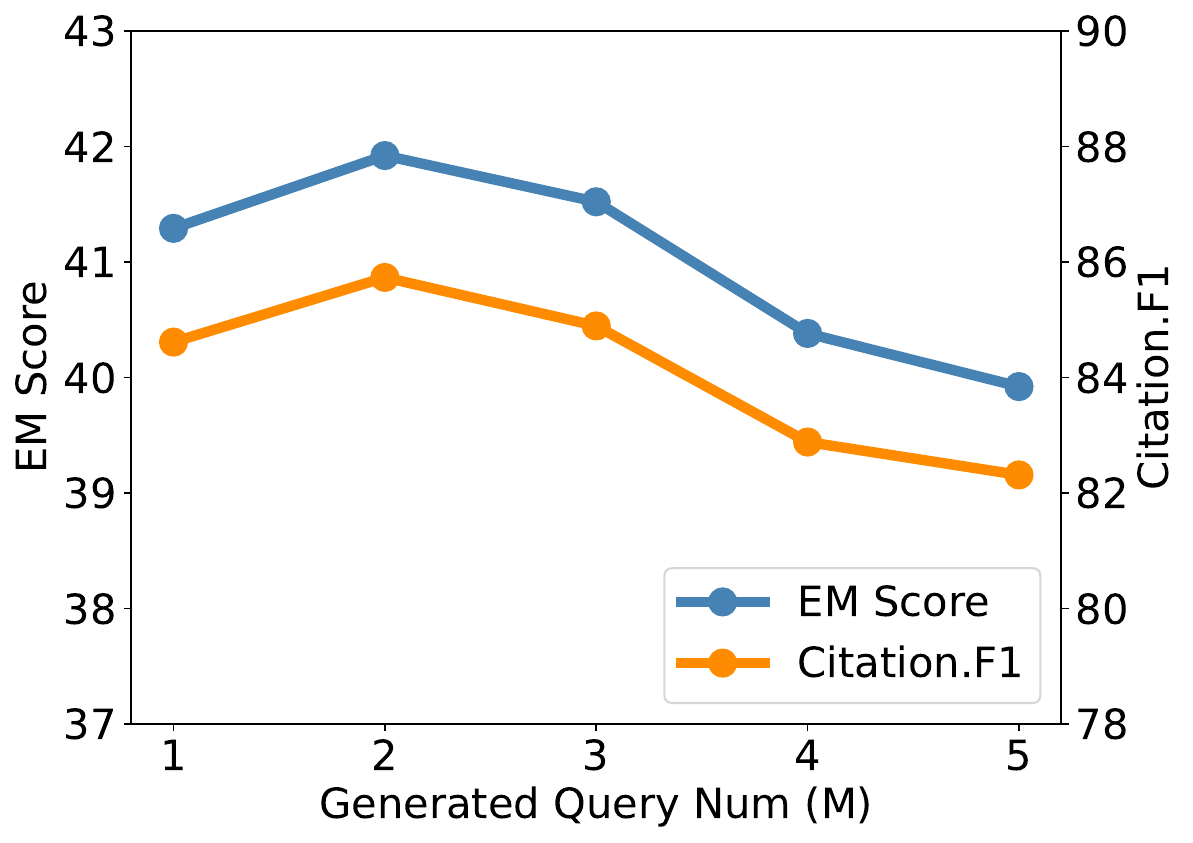}
    }
    \subfigure[Retrieval Num]{
    \includegraphics[width=0.22\textwidth]{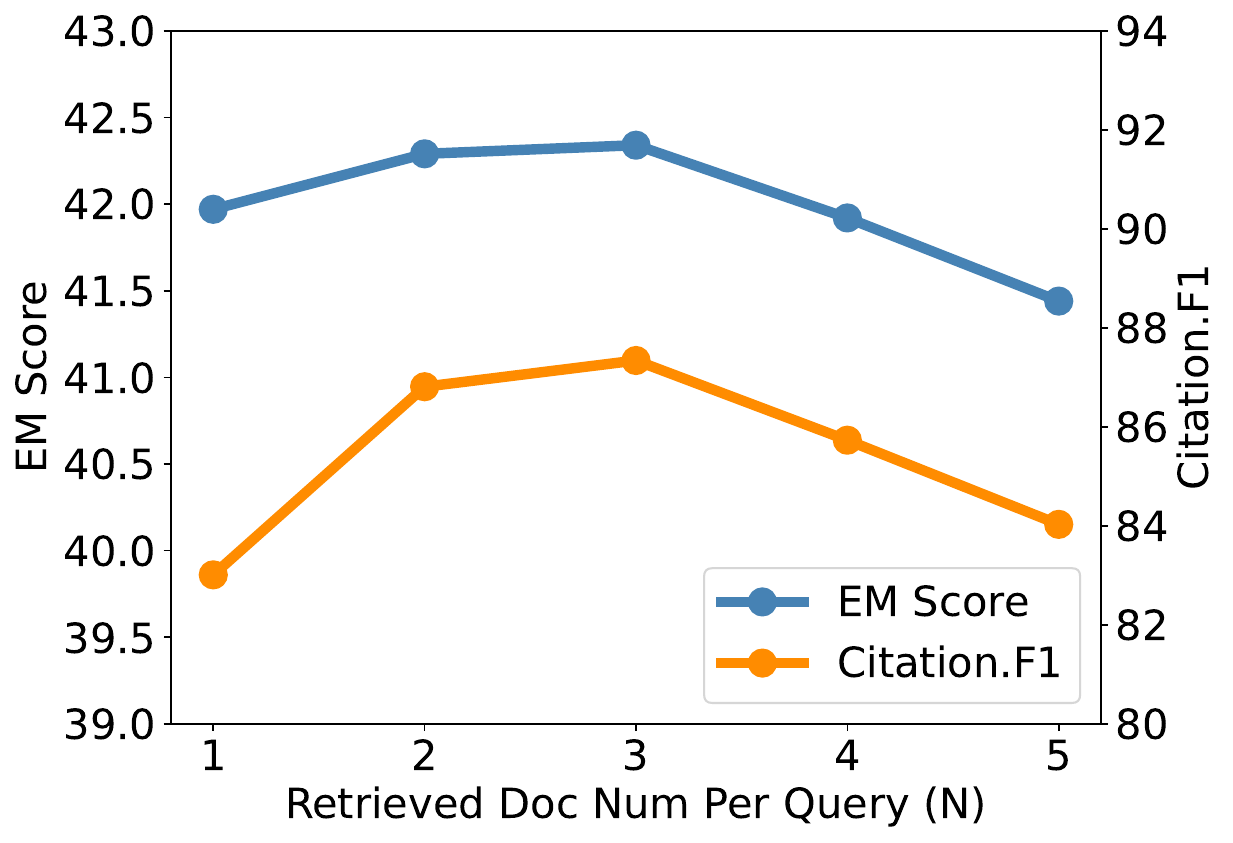}
    }
    \caption{The performance change over different hyper-parameters on ASQA.}
    \label{fig:hyper}
\end{figure}

\subsection{Analysis}
In this section, we conduct analytical experiments of our method using \texttt{Vicuna-13B} as the default LLM, unless specified otherwise.

\paragraph{Ablation Study.}
We assess the impact of each component in our model using the 2WikiMultihopQA dataset. By removing components one by one, we observe their individual contributions to performance. ``w/o Verifier'' excludes the two-tier verifier during generation, ``w/o Memory'' removes the evolving memory system, ``w/o Simplifier'' removes the citation simplifier and ``w/o Diverse QG'' replaces diverse query generation with direct retrieval using the claim as the query.

Results in \cref{tab:ablation} show that removing any component decreases performance, highlighting their importance.
Notably, the removal of the verifier results in the most significant drop in performance, as it potentially leads to the generation of claims with hallucinations or factual inaccuracies.
Omitting the simplifier has a less pronounced effect on correctness, as all claims are still verified. However, it does lead to unnecessary citations, which reduces the precision of citation quality.
Removing the memory component also results in a decline in performance, affecting both correctness and citation quality. This is primarily due to the lack of supporting evidence for the constantly changing topic.
Lastly, removing diverse QG limits the model's ability to retrieve a broader range of relevant documents, leading to a degradation in performance.

\paragraph{Performance over Max Trials.}
We examine the impact of various max trials $T$ on \ours{}'s performance using 2WikiMultihopQA dataset.
As illustrated in \cref{tab:max_trials}, we observed that increasing the value of $T$ correlates with improved performance in terms of correctness and citation quality.
This improvement is reasonable since a higher $T$ allows the model more attempts to generate a claim that passes the verification process, increasing the chances of generating accurate and well-supported claims.
However, it's important to acknowledge that a higher $T$ also leads to larger token consumption, indicating the need to adjust $T$ to balance between effectiveness and computational cost.

\paragraph{Retrieval Analysis.} \begin{table}[t]
\centering
\resizebox{\linewidth}{!}{
\begin{tabular}{ccccccc}
\toprule
& \multicolumn{2}{c}{\tf{Correct}}  & \multicolumn{3}{c}{\tf{Citation} }  &\tf{Cost} \\
\cmidrule(lr){2-3} \cmidrule(lr){4-6} \cmidrule(lr){7-7} 
\textbf{$T$} & \textbf{EM} & \textbf{F1} & \textbf{Rec} & \textbf{Prec} & \textbf{F1}  & \textbf{Trials} \\
\midrule
1  & 22.00 & 19.56 & 41.20 & 34.07 & 37.29 & 1.82  \\
2  & 22.60 & 20.22 & 47.82 & 41.95 & 44.69 & 2.15  \\
3  & 22.80 & 20.63 & 49.95 & 44.03 & 46.80 & 2.46  \\
4  & 25.00 & 22.66 & 51.37 & 45.70 & 48.36 & 2.72  \\
5  & 25.60 & 23.27 & 55.36 & 49.59 & 52.31 & 2.92  \\
\bottomrule
\end{tabular}
}
\caption{Performance of \ours{} with respect to the max trials $T$ on 2WikiMultihopQA, where the ``Trials'' represent the average iteration it takes to complete a claim.}
\label{tab:max_trials}
\end{table}

We examine the impact of retrieval parameters on \ours{}'s performance using ASQA dataset, focusing on the number of generated queries $M$ and the number of retrieved documents per query $N$.
As shown in \cref{fig:hyper}, we observe a consistent trend for both parameters. 
Initially, increasing $M$ and $N$ enhances both correctness and citation quality, which can be attributed to the fact that a larger pool of documents offers a broader knowledge scope and provides more citation options for the LLM.
However, continually increasing them beyond a certain threshold results in a decline in performance, which is mainly because an excessively large document pool can introduce noisy information, negatively impacting both claim generation and citation generation processes.

\begin{figure}[]
    \centering
    \includegraphics[width=\linewidth]{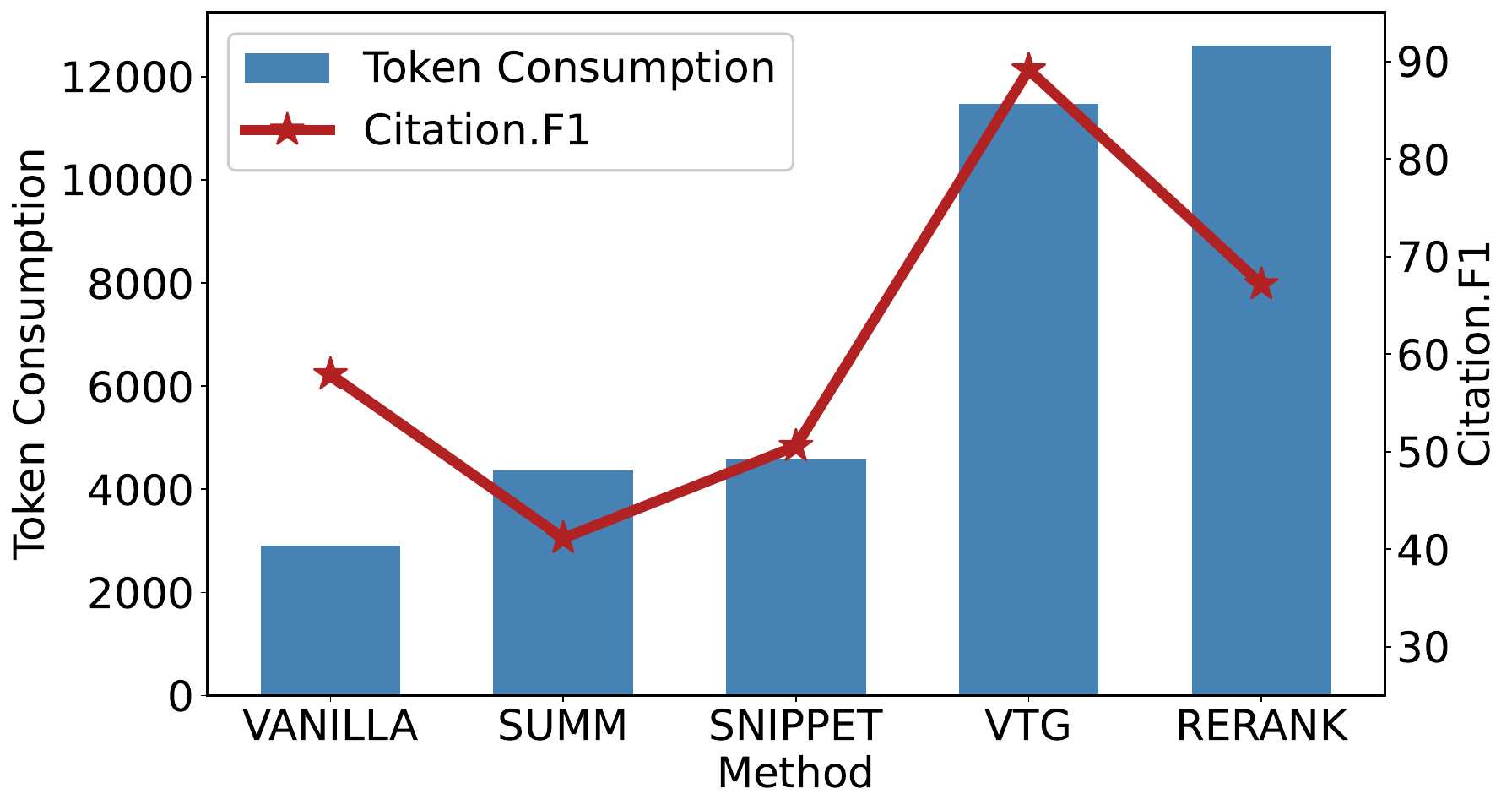}
    \caption{Token Consumption Comparison on NQ.}
    \label{fig:token_consume}
\end{figure}

\paragraph{Token Consumption Analysis.}
We analyze token consumption across various methods on the NQ dataset using \texttt{Text-Davinci-003} as the LLM.
As shown in \cref{fig:token_consume}, \vani{} achieves the lowest token cost due to its single API request.
After introducing more documents, \summ{}, \snippet{} incur higher token costs but with lower citation quality as simplified documents challenge LLMs' citation generation.
\rerank{} produces better citation quality than other baselines but incurs the highest token cost due to its strategy of multiple sampling.
In contrast, our method \ours{} demonstrates a lower token cost than \rerank{} while significantly improving citation quality, highlighting the superiority of our approach.
Importantly, users can adjust \ours{}'s token cost with the max trial parameter $T$ to balance performance and computational cost.
\section{Related Work}
\subsection{Retrieval-augmented LLMs.}
Retrieval-augmented LLMs aim to provide extra documents to the LLM, which has been proven useful in many knowledge-intensive tasks.
Among the existing studies, \citet{shi2023replug, wang2023learning, zhang2023merging, yu2023chain, yu2023augmentation}  propose to retrieve only once at the beginning.
Other works~\cite{qian2023optimizing, yu2023improving} propose to retrieve multiple times during generation, which offers the flexibility of when and what to search.
For example, 
\citet{jiang2023active} propose to retrieve when the generation contains low confidence tokens.
\citet{ram2023context} propose to refresh the retrieved document every n token, which is demonstrated to be more effective than retrieving only once.
\citet{wang2023self,asai2023self,zhao2023thrust} propose to retrieve only when the LLMs need to.
Among the existing studies, retrieve on-the-fly methods are the closest to ours. However, these methods do not provide referenced documents for the generated sentences, potentially reducing the reliability of the generated content.

\subsection{Verifiable Text Generation}
Verifiable Text Generation aims to generate content with supporting documents, which has been attracting attention in recent years.
For example, 
\citet{liu2023webglm,qin2023webcpm,nakano2112webgpt} focus on training LLMs to browse web pages and answer questions with evidence.
\citet{gao2023rarr} introduced the research-and-revision (RARR) method for retrieving evidence for LLM outputs.
\citet{li2023towards} incorporated knowledge graphs as an evidence source.
Other works mainly focus on evaluation~\cite{liu2023evaluating}.
For example,
\citet{rashkin2023measuring} propose \textit{Attributable to Identified Sources} (AIS) for human evaluation. 
\citet{gao2022rarr} define auto-AIS to approximate human AIS judgments. 
\citet{gao2023enabling} propose ALCE  to automate the evaluation of the citation quality.
\citet{min2023factscore} proposed FactScore for evaluating the verifiability of generated facts.
Although these methods achieve promising results, they do not effectively address the focus-shifting phenomenon and fail to capture the complex relationship between claims and citations.
In this paper, we propose to maintain an evolving memory and two-tier verification to deal with the above-mentioned issues.
\section{Conclusion}
In this paper, we introduce \ours{}, a novel framework tailored to address the challenges of verifiable text generation. 
Central to its design is an evolving long short-term memory, which adaptively keeps both valuable documents and up-to-date documents.
The two-tier verifier coupled with an evidence finder facilitates a deeper analysis and reflection on the relationship between claims and citations.
Through the integration of active retrieval mechanisms and diverse query generation, \ours{} skillfully enhances both the precision and breadth of the document retrieval process.
Extensive experiments on five datasets across three knowledge-intensive tasks verify the effectiveness of our method.
\section*{Limitations}
In this work, we propose a novel frame \ours{} for verifiable text generation.
The limitations of the proposed method are as follows:
(1) the computational cost of \ours{} is relatively high due to the need for multiple API calls and frequent verification. This may restrict its applicability in resource-intensive scenarios or systems with limited computational resources;
(2) the effectiveness of our verification process is constrained by the precision of the NLI models. In instances where the NLI model's accuracy is suboptimal, there is a risk of incorporating erroneous information into the method, potentially compromising the verifiability of the generated text.

As for future work, we plan to mitigate the computational cost of the method by developing more efficient pipelines. Moreover, we aim to reduce our approach's reliance on NLI models, thereby enhancing the overall robustness of our framework.

\section*{Acknowledgement}
This work is supported in part by Ucap Cloud and the State Key Laboratory of General Artificial Intelligence.
\section*{Ethics Statement}
This work was conducted in rigorous compliance with the ACL Ethics Policy. All datasets and large language models (LLMs) used for evaluation are publicly available. Furthermore, our work strives to improve the verifiability of LLM outputs, which could potentially broaden the application scenarios of LLMs. We do not foresee any form of negative ethical impact induced by our work.

\bibliography{custom}
\clearpage
\appendix

\begin{table*}[tb]
\centering
\resizebox{\linewidth}{!}{
\begin{tabular}{lccccc}
\toprule
\textbf{Settings} & \textbf{2WikiMultihopQA} & \textbf{ASQA} & \textbf{ELI5} & \textbf{NQ} & \textbf{WebQ} \\
 & \cite{2wikimultihopqa-ho-2020}  & \cite{asqa-stelmakh-2022} & \cite{fan-etal-2019-eli5} & \cite{kwiatkowski2019natural} & \cite{berant2013semantic} \\
\midrule
& \multicolumn{5}{c}{\emph{Dataset statistics}} \\
Task & Multihop QA & Long-form QA & Long-form QA & Open-domain QA & Open-domain QA  \\
\#Examples (Vicuna) & 500 & 500 & 500 & 500 & 500 \\
\#Examples (Davinci) & 200 & 200 & 200 & 200 & 200 \\
\midrule
& \multicolumn{5}{c}{\emph{Evaluation settings}} \\
Correctness Metrics & EM, Token-level F1, Pre, and Rec & EM, Disambig-F1, ROUGE, DR & Claim & EM, F1 & EM, F1 \\
Citation Metrics & \multicolumn{5}{c}{Prec, Rec, F1}  \\
\midrule
& \multicolumn{5}{c}{\emph{Retrieval settings}} \\
Corpus & Wikipedia & Wikipedia & Wikipedia & Wikipedia & Wikipedia \\
Retriever & DPR & DPR  & DPR& DPR& DPR  \\
\bottomrule
\end{tabular}
}
\caption{Statistics and experimental settings of different tasks/datasets.}
\label{tab:setting}
\end{table*}
\begin{table*}[t]
\centering
\small
\newcolumntype{C}[1]{>{\centering\arraybackslash}m{#1}}
\begin{tabular}{l|C{1.6cm}|C{1.6cm}|C{1.6cm}|C{1.6cm}|C{1.6cm}}
\toprule
\textbf{Parameter} & \textbf{MultihopQA} & \textbf{ASQA} & \textbf{ELI5}  & \textbf{NQ} & \textbf{WebQ}   \\
\midrule
Maximum trials $T$   & 5  &  2  &  3 & 3  &  3 \\
Generated queries number $M$    &  2 &  4  &  2 &  2 &  4  \\
Initially retrieved documents number $k$  &  5 & 5   & 5  &  5 &  5 \\
Retrieved documents number  per query $N$   &  4 &  2  &  2 &  3 & 2   \\
\bottomrule
\end{tabular}
\caption{Hyper-parameters for \ours{} on \texttt{Text-Davinci-003}, where MultihopQA refers to 2WikiMultihopQA.}
\label{tab:hyps_davinci}
\end{table*}

\begin{table*}[t]
\centering
\small
\newcolumntype{C}[1]{>{\centering\arraybackslash}m{#1}}
\begin{tabular}{l|C{1.6cm}|C{1.6cm}|C{1.6cm}|C{1.6cm}|C{1.6cm}}
\toprule
\textbf{Parameter} & \textbf{MultihopQA} & \textbf{ASQA} & \textbf{ELI5}  & \textbf{NQ} & \textbf{WebQ}   \\
\midrule
Maximum trials $T$   & 5  &  3  & 5 & 3 & 3  \\
Generated queries number $M$    &  4 &  2  &  2 & 3 &  3 \\
Initially retrieved documents number $k$  &  5 & 5   & 5  &  5 &  5 \\
Retrieved documents number  per query $N$   &  2 &  4  & 4  & 3 &  3 \\
\bottomrule
\end{tabular}
\caption{Hyper-parameters for \ours{} on \texttt{Vicuna-13B-v1.5-16k}, where MultihopQA refers to 2WikiMultihopQA.}
\label{tab:hyps_vicuna}
\end{table*}


\begin{algorithm}[t]
\small
\begin{algorithmic}[1]
 \renewcommand{\COMMENT}[2][.1\linewidth]{\leavevmode\makebox[#1][l]{\#~#2}}
\caption{The pipeline of \ours{}}
\label{alg:VTG}
\renewcommand{\algorithmicrequire}{\textbf{Input:}}
\renewcommand{\algorithmicensure}{\textbf{Output:}}

\REQUIRE Question $q$, document pool $\mathbb{D}$, the Generators, the Verifiers, the Citation Simplifier, the Evidence Finder, the maximum trials $T$, the retriever $R$, the number of initially retrieved documents $k$, the number of generated queries $M$, the number of documents retrieved per query $N$
\ENSURE Output with citations $O$
\STATE $t \gets 0$
\STATE $O \gets \{\}$
\STATE $D_S \gets \{\}$
\STATE $D_L \gets R(q,\mathbb{D}, k)$

\WHILE{TRUE}
    \STATE $s \gets \operatorname{ClaimGenerator}(O, D_S \cup D_L)$ 
    \STATE $C \gets \operatorname{CitationGenerator}(s, D_S \cup D_L)$ 
    \IF{ $s$ is \texttt{<EOS>} }
        \STATE break
    \ENDIF
    \IF{ $\operatorname{GenerationVerifier}(s,C) \to$ TRUE }
        \STATE $C \gets \operatorname{CitationSimplifier}(s,C)$
        \STATE $O \gets O \cup \{s,C\}$ 
        \STATE $D_L \gets D_L \cup C$
        \STATE $t \gets 0$
    \ELSIF{$\operatorname{MemoryVerifier}(s,D_S \cup D_L) \to$ TRUE}
        \STATE $C \gets \operatorname{CitationSimplifier}(s,D_S \cup D_L)$
        \STATE $O \gets O \cup \{s,C\}$
        \STATE $D_L \gets D_L \cup C$
        \STATE $t \gets 0$
    \ELSIF{$t > T$}
        \STATE $O \gets O \cup \{s,C\}$
        \STATE $t \gets 0$
    \ELSE
        \STATE $D_S \gets \operatorname{EvidenceFinder}(s,M,N)$
        \STATE $t \gets t+1$
    \ENDIF
    \ENDWHILE
    \RETURN $O$
\end{algorithmic} 
\end{algorithm}

\section{Baselines}\label{baselines}
For an equitable comparison, we have selected four best-performing baseline methodologies as proposed in ALCE \cite{gao2023enabling}, which include \textsc{\pvani{}}, \textsc{\summ{}}, \textsc{\snippet{}} and \textsc{\rerank{}}.
Each of these methods incorporates multiple demonstrations within the initial prompt to facilitate the process of generating responses.
Following ALCE \cite{gao2023enabling}, we set $k = 5$ and $K = 10$ in our experiment.

\paragraph{\textsc{\pvani{}}.}
This configuration involves providing the LLM with the top-$k$ ranked documents. The LLM is then tasked with generating responses that appropriately include citations.

\paragraph{\textsc{\psumm{}}.}
In this approach, the LLM is required to synthesize relevant information from the top-$K$ ranked documents. After summarizing these documents, the condensed text is integrated into the prompt. The LLM is then instructed to create texts that incorporate citations, drawing from this summarized content.

\paragraph{\textsc{\psnippet{}}.}
In this setup, the LLM is instructed to extract relevant snippets from the top-$K$ ranked documents. These concise documents are subsequently utilized in the prompt, with the aim for the LLM to create text that includes citations, drawing from these brief extracts.

\paragraph{\textsc{\prerank{}}.}
This methodology entails a two-stage process. Initially, the LLM generates four distinct responses based on the top-$k$ ranked documents using high temperature. Thereafter, each response undergoes an evaluation for citation recall. The response with the highest citation recall score is then chosen as the final output.

\paragraph{\textsc{\pposthoc{}}:}
The LLM first generates answers then cites the most relevant passage from the top-100 retrieved documents for each statement.

\paragraph{\textsc{\prefinecite{}}:}
The LLM first produces answers with citations, then uses NLI to remove unnecessary citations, refining the citation set.

\paragraph{\textsc{\pvericite{}}:}
The LLM first generates sentences with citations and then uses an NLI method to ensure the citation set conclusively supports the sentence. If not, it finds and cites the best matching passage from the top-100 retrieved documents using GTR.

\paragraph{\textsc{\pverirefine{}}:}
Integrates RefineCite's refinement and VeriCite's verification processes, ensuring citations are both necessary and fully supportive, optimizing citation accuracy and relevance

\begin{table*}[tb]
\centering
\resizebox{\linewidth}{!}{
\begin{tabular}{lcccccccccccccccccccccc}
\toprule
\textbf{Datasets} & \multicolumn{8}{c}{\textbf{Wikihop}} & 
\multicolumn{7}{c}{\textbf{WebQ}}  & 
\multicolumn{7}{c}{\textbf{NQ}}  \\
 \cmidrule(lr){2-9}  \cmidrule(lr){10-16}  \cmidrule(lr){17-23} 
& \multicolumn{2}{c}{\tf{Correct}}  & \multicolumn{3}{c}{\tf{ALCE.Citation} } & \multicolumn{3}{c}{\tf{LLM.Citation} } &\tf{Correct} & \multicolumn{3}{c}{\tf{ALCE.Citation} } & \multicolumn{3}{c}{\tf{LLM.Citation} } & \tf{Correct} & \multicolumn{3}{c}{\tf{ALCE.Citation}} & \multicolumn{3}{c}{\tf{LLM.Citation} } \\

\cmidrule(lr){2-3} \cmidrule(lr){4-6}  \cmidrule(lr){7-9}  \cmidrule(lr){10-10}  \cmidrule(lr){11-13}   \cmidrule(lr){14-16} \cmidrule(lr){17-17} \cmidrule(lr){18-20}   \cmidrule(lr){21-23} 

\textbf{Metrics} & \textbf{EM} & \textbf{F1} & \textbf{Rec} & \textbf{Prec} & \textbf{F1} & \textbf{Rec} & \textbf{Prec} & \textbf{F1} & \textbf{EM} & \textbf{Rec} & \textbf{Prec} & \textbf{F1} & \textbf{Rec} & \textbf{Prec} & \textbf{F1} & \textbf{EM} & \textbf{Rec} & \textbf{Prec} & \textbf{F1} & \textbf{Rec} & \textbf{Prec} & \textbf{F1} \\
\midrule
\headercolor
\multicolumn{23}{c}{\textbf{Vicuna-13B}} \\
\textsc{\pvani{}} &   23.40 & 21.98 & 29.55 & 22.25 & 25.39 & 41.59 & 35.04 & 38.03 & 55.80 & 67.66 & 60.66 & 63.97 & 67.50 & 67.83 & 67.67 & 54.80 & 71.39 & 61.71 & 66.20 & 77.46 & 63.94 & 70.05  \\
\summ{} &  23.20 & 20.00 & 30.89 & 28.43 & 29.61 & 37.66 & 39.12 & 38.37 & 58.00 & 70.51 & 62.07 & 66.02 & 68.23 & 66.57 & 67.39 & 57.00 & 51.55 & 52.21 & 51.88 & 56.76 & 62.86 & 59.66   \\
\snippet{} & 21.80 & 20.05 & 25.18 & 21.95 & 23.45 & 33.67 & 29.75 & 31.59 & 58.40 & 53.44 & 49.15 & 51.21 & 68.46 & 69.57 & 69.01 & 57.20 & 43.56 & 41.43 & 42.47 & 57.56 & 59.57 & 58.55  \\
\rerank{} &  22.60 & 21.13 & 47.03 & 47.53 & 47.28 & 54.33 & 53.73 & 54.03 & 56.40 & 89.93 & 76.33 & 82.57 & 88.20 & 67.77 & 76.64 & 56.20 & 83.56 & 73.57 & 78.25 & 81.66 & 73.49 & 77.36   \\

\posthoc{} &  22.20 & 15.01 & 8.34 & 6.90 & 7.55 & 22.59 & 15.76 & 18.57 & 58.60 & 46.26 & 32.73 & 38.34 & 65.66 & 61.11 & 63.30 & 46.60 & 35.21 & 24.40 & 28.83 & 52.88 & 50.98 & 51.91  \\
\refinecite{} & 24.20 & 22.15 & 48.00 & 39.50 & 43.34 & 52.61 & 47.12 & 49.72 & 55.40 & 78.04 & 72.55 & 75.19 & 84.44 & 79.72 & 82.01 & 55.20 & 65.05 & 61.80 & 63.39 & 75.47 & 77.00 & 76.23  \\
\vericite{} & 22.89 & 21.29 & 14.69 & 8.21 & 10.53 & 41.36 & 24.77 & 30.98 & 57.20 & 80.97 & 54.41 & 65.08 & 78.97 & 69.07 & 73.69 & 54.80 & 75.81 & 54.17 & 63.19 & 83.81 & 65.83 & 73.74    \\
\verirefine{} &  24.61 & 23.06 & 15.90 & 11.30 & 13.21 & 47.57 & 32.97 & 38.94 & 54.40 & 82.33 & 64.67 & 72.44 & 81.33 & 75.00 & 78.04 & 55.60 & 71.34 & 58.35 & 64.19 & 81.34 & 67.52 & 73.79   \\
\ours{} &\textbf{ 25.60} & \textbf{23.27} & \textbf{55.36} & \textbf{49.59} & \textbf{52.32} & \textbf{62.76} &\textbf{ 54.69} & \textbf{58.45} & \textbf{60.00} & \textbf{92.16} &\textbf{ 86.51} & \textbf{89.25} & \textbf{89.43} & \textbf{81.38} & \textbf{85.21} &\textbf{ 58.00 }& \textbf{88.69} & \textbf{82.02} &\textbf{ 85.22} & \textbf{86.35} & \textbf{78.06 }& \textbf{82.00}   \\
\midrule
\headercolor
\multicolumn{23}{c}{\textbf{Text-Davinci-003}} \\
\vani{} &  33.00 & 33.01 & 40.46 & 28.30 & 33.30 & 59.07 & 43.00 & 49.77 & 67.50 & 63.78 & 58.97 & 61.28 & 71.18 & 66.52 & 68.77 & 62.50 & 60.48 & 55.56 & 57.92 & 66.45 & 61.59 & 63.93   \\
\summ{} &  30.00 & 30.63 & 9.39 & 12.19 & 10.61 & 23.19 & 24.64 & 23.89 & 67.50 & 60.06 & 47.62 & 53.12 & 68.33 & 56.65 & 61.95 & 62.50 & 44.23 & 38.45 & 41.14 & 55.36 & 49.20 & 52.10    \\
\snippet{} &  32.00 & 30.13 & 13.86 & 18.49 & 15.84 & 37.36 & 38.99 & 38.16 & 67.00 & 65.41 & 52.32 & 58.14 & 71.81 & 68.07 & 69.89 & 62.00 & 54.72 & 46.99 & 50.56 & 73.05 & 69.55 & 71.25   \\
\rerank{} &  32.67 & 33.09 & 56.13 & 45.22 & 50.09 & 63.43 & 46.32 & 53.54 & 67.00 & 73.72 & 64.90 & 69.03 & 78.12 & 70.00 & 73.84 & 61.50 & 71.30 & 63.44 & 67.14 & 79.03 & 66.37 & 72.15  \\
\ours{} &  \textbf{41.50} & \textbf{40.19} & \textbf{63.89} & \textbf{57.65} & \textbf{60.61} & \textbf{70.47} & \textbf{59.13} & \textbf{64.30} & \textbf{68.00} & \textbf{93.00} & \textbf{88.72} & \textbf{90.81} & \textbf{90.70} & \textbf{87.52} & \textbf{89.08} & \textbf{63.00} & \textbf{91.85} & \textbf{86.59} & \textbf{89.14} & \textbf{84.92} & \textbf{73.20} & \textbf{78.63}   \\
\bottomrule
\end{tabular}
}
\caption{Comparisons between \ours{} and baselines on Multi-hop QA task and Open-domain QA task.}
\label{tab:main_result_3}
\end{table*}

\begin{table*}[tb]
\centering
\resizebox{\linewidth}{!}{
\begin{tabular}{lcccccccccccccccccccc}
\toprule
\textbf{Datasets} & \multicolumn{10}{c}{\textbf{ASQA}} & 
\multicolumn{7}{c}{\textbf{ELI5}} & 
\multicolumn{2}{c}{\textbf{Overall}}  \\
\cmidrule(lr){2-11} \cmidrule(lr){12-18} \cmidrule(lr){19-20}
& \multicolumn{4}{c}{\tf{Correct}}  & \multicolumn{3}{c}{\tf{ALCE.Citation} } & \multicolumn{3}{c}{\tf{LLM.Citation} } &\tf{Correct} & \multicolumn{3}{c}{\tf{ALCE.Citation}} & \multicolumn{3}{c}{\tf{LLM.Citation} } & \tf{Correct} & \tf{Citation}  \\

\cmidrule(lr){2-5} \cmidrule(lr){6-8} \cmidrule(lr){9-11} \cmidrule(lr){12-12} \cmidrule(lr){13-15} \cmidrule(lr){16-18}  \cmidrule(lr){19-19} \cmidrule(lr){20-20}
\textbf{Metrics} & \textbf{EM} & \textbf{D-F1} & \textbf{R-L} & \textbf{DR} & \textbf{Rec} & \textbf{Prec} & \textbf{F1} & \textbf{Rec} & \textbf{Prec} & \textbf{F1} & \textbf{Claim} & \textbf{Rec} & \textbf{Prec} & \textbf{F1}  &  \textbf{Rec} & \textbf{Prec} & \textbf{F1} &\textbf{EM}  & \textbf{F1} \\
\midrule
\headercolor
\multicolumn{20}{c}{\textbf{Vicuna-13B}} \\
\vani{} &  32.00 & 27.52 & 33.53 & 30.53 & 72.78 & 62.09 & 67.01 & 73.28 & 66.59 & 69.78 & 12.20 & 59.79 & 48.26 & 53.41 & 81.46 & 76.89 & 79.11 & 35.64 & 60.06   \\
\summ{}  &  41.71 & 28.95 & 37.18 & 33.07 & 62.15 & 59.60 & 60.85 & 68.95 & 70.19 & 69.56 & 14.20 & 60.13 & 52.42 & 56.01 & 77.87 & 72.42 & 75.05 & 38.82 & 57.44   \\
\snippet{} &  39.22 & 27.01 & 35.65 & 31.33 & 46.23 & 47.04 & 46.63 & 56.55 & 63.03 & 59.61 & 14.33 & 31.47 & 32.72 & 32.08 & 46.69 & 49.32 & 47.97 & 38.19 & 46.26    \\
\rerank{} &  37.14 & 28.21 & 32.18 & 30.20 & 88.29 & 75.74 & 81.53 & 88.29 & 75.74 & 81.53 & 11.67 & 73.80 & 61.12 & 66.86 & 84.57 & 77.09 & 80.65 & 36.80 & 72.67   \\
\posthoc{} & 25.55 & 22.24 & 35.36 & 28.80 & 38.54 & 25.69 & 30.83 & 47.14 & 38.14 & 42.16 & 14.47 & 26.90 & 17.02 & 20.85 & 62.15 & 52.20 & 56.74 & 33.48 & 35.91     \\
\refinecite{}  &  36.30 & 28.24 & 35.35 & 31.79 & 77.80 & 60.50 & 68.07 & 69.80 & 61.17 & 65.20 & 11.60 & 26.90 & 59.05 & 36.96 & 32.15 & 67.46 & 43.54 & 36.54 & 60.36     \\
\vericite{}  &  34.36 & 28.86 & 36.16 & 32.51 & 76.15 & 72.67 & 74.37 & 77.32 & 75.83 & 76.57 & 13.73 & 64.93 & 40.14 & 49.61 & 77.93 & 62.48 & 69.35 & 36.60 & 58.71   \\
\verirefine{} &  34.36 & 28.86 & 35.40 & 32.13 & 80.80 & 59.28 & 68.39 & 76.80 & 64.95 & 70.38 & 11.27 & 61.57 & 37.59 & 46.68 & 64.57 & 51.26 & 57.15 & 36.05 & 58.32   \\
\ours{} & \textbf{41.92} & \textbf{30.53} & \textbf{37.87} & \textbf{34.20} & \textbf{89.15} & \textbf{82.57} & \textbf{85.73} & \textbf{89.15} & \textbf{82.57} & \textbf{85.73} & \textbf{14.73} & \textbf{81.50} & \textbf{72.16} & \textbf{76.55} & \textbf{87.60} & \textbf{84.46} & \textbf{86.00}& \textbf{40.05} & \textbf{78.65}   \\
\midrule
\headercolor
\multicolumn{20}{c}{\textbf{Text-Davinci-003}} \\
\vani{} &  40.25 & 31.47 & 35.81 & 33.64 & 58.13 & 55.17 & 56.61 & 58.13 & 55.17 & 56.61 & 13.43 & 58.66 & 47.40 & 52.43 & 58.66 & 47.40 & 52.43 & 43.34 & 55.31   \\
\summ{} &  41.33 & 28.91 & 37.21 & 33.06 & 48.31 & 40.68 & 44.17 & 50.48 & 44.44 & 47.27 & 11.50 & 39.43 & 31.81 & 35.21 & 52.27 & 48.47 & 50.30 & 42.57 & 41.98  \\
\snippet{} &  39.60 & 30.11 & 38.35 & 34.23 & 53.14 & 43.19 & 47.65 & 59.31 & 52.05 & 55.44 & 13.67 & 45.29 & 37.23 & 40.87 & 62.39 & 55.19 & 58.57 & 42.85 & 50.64   \\
\rerank{} &  39.55 & 29.94 & 39.38 & 34.66 & 75.83 & 69.81 & 72.70 & 76.41 & 70.01 & 73.07 & 14.76 & 76.21 & 61.67 & 68.17 & 86.98 & 77.64 & 82.04 & 43.10 & 68.18   \\
\ours{} & \textbf{41.53} & \textbf{31.64} & \textbf{39.45} & \textbf{35.55} & \textbf{86.70} & \textbf{79.95} & \textbf{83.19} & \textbf{89.10} & \textbf{79.84} & \textbf{84.22} & \textbf{16.67} & \textbf{82.63} & \textbf{71.56} & \textbf{76.70} & \textbf{87.94} & \textbf{81.79} & \textbf{84.75} & \textbf{46.14} & \textbf{80.14}   \\
\bottomrule
\end{tabular}
}
\caption{Comparisons between \ours{} and baselines on Long-form QA task and overall performance.}
\label{tab:main_result_4}
\end{table*}

The complete experimental result is shown in \cref{tab:main_result_3} and \cref{tab:main_result_4}.

\section{Datasets and Settings}\label{ap:setting}
Datasets and experimental settings are summarized in \cref{tab:setting}.

\paragraph{Citation Recall.}
Citation recall for each claim in the model's response is computed individually as either 0 or 1 and then averaged across all claims in the response. 
A claim's citation recall is 1 if at least one citation exists and the concatenated citations entail the claim according to an NLI model, which outputs 1 for entailment.

\paragraph{Citation Precision.} 
Citation precision for each citation in the model's response is computed individually as either 0 or 1 and then averaged across all citations in the response.
The precision score for a citation is 1 if the associated claim has a citation recall of 1 and the citation is not irrelevant; otherwise, it's 0.
A citation is deemed irrelevant if: (a) the citation alone cannot substantiate the claim, and (b) omitting the citation doesn’t impact the remaining citations’ ability to support the claim.

\paragraph{LLM Evaluation}
To ensure a fair comparison of citation quality, we instruct Qwen-MAX to evaluate model generations. Specifically, we assess the quality of the citations in two ways: (1) Citation Recall: The large language model (LLM) is given a sentence and all the passages that the sentence cited, and is asked to judge whether the passages fully support the sentence; (2) Citation Precision: Given a sentence and one of its citations, the LLM is asked to judge whether the citation “fully supports” or “does not support” the sentence. Each citation receives a precision score of 1 if the output sentence has a citation recall of 1 and this citation is “fully support.”

\section{Focus Shifting Phenomenon} 
\begin{figure*}[t]
    \centering
    \subfigure[Wikihop]{
    \includegraphics[width=0.18\textwidth]{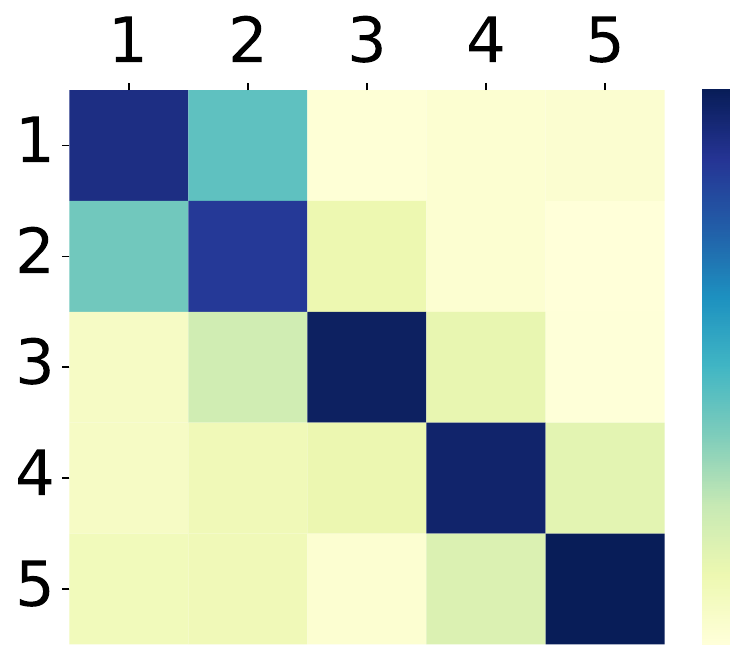}
    }
    \subfigure[WebQ]{
    \includegraphics[width=0.18\textwidth]{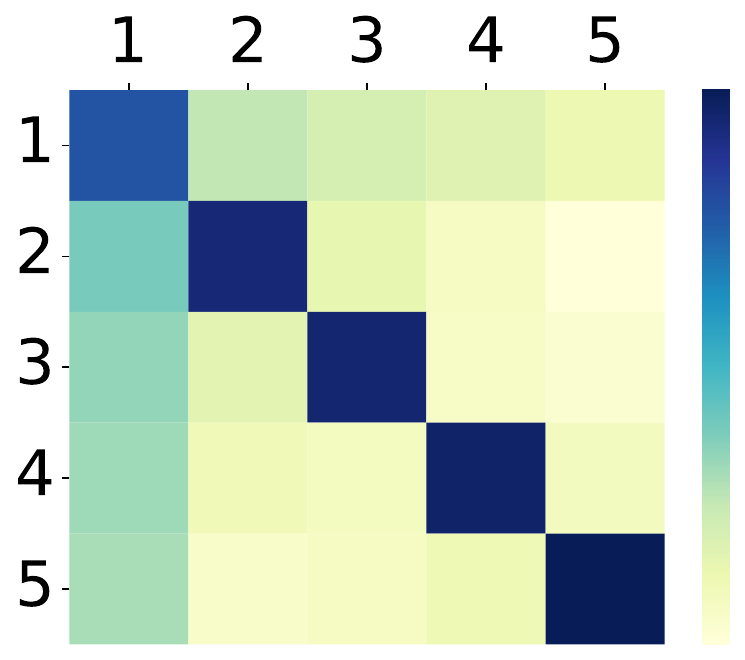}
    }
    \subfigure[NQ]{
    \includegraphics[width=0.18\textwidth]{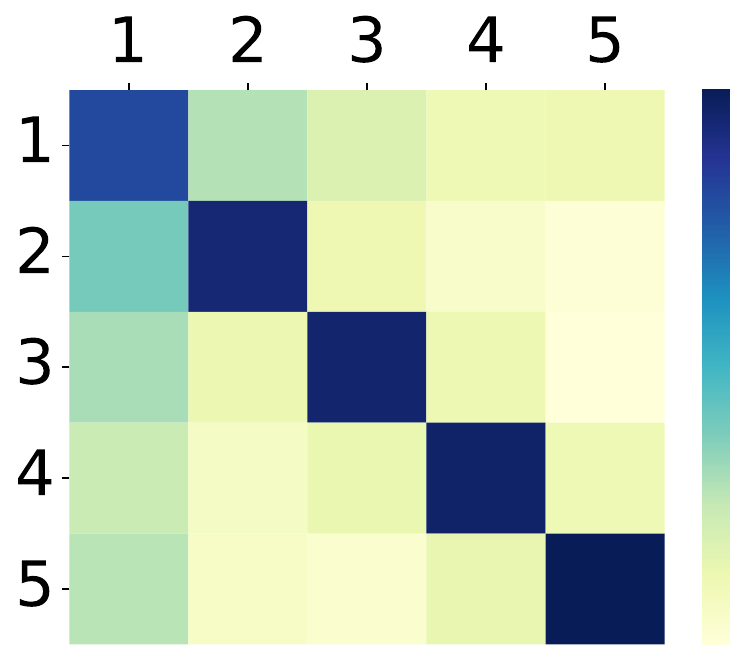}
    }
    \subfigure[ASQA]{
    \includegraphics[width=0.18\textwidth]{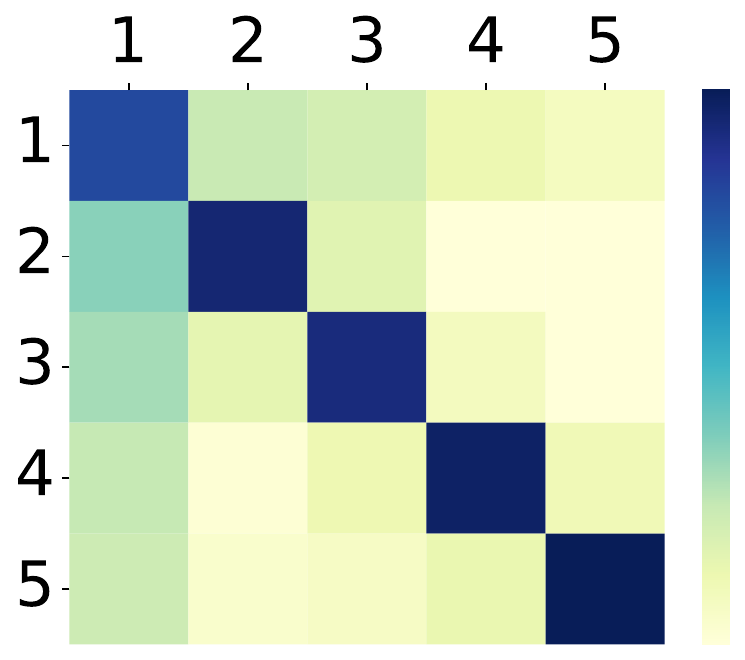}
    }
    \subfigure[ELI5]{
    \includegraphics[width=0.18\textwidth]{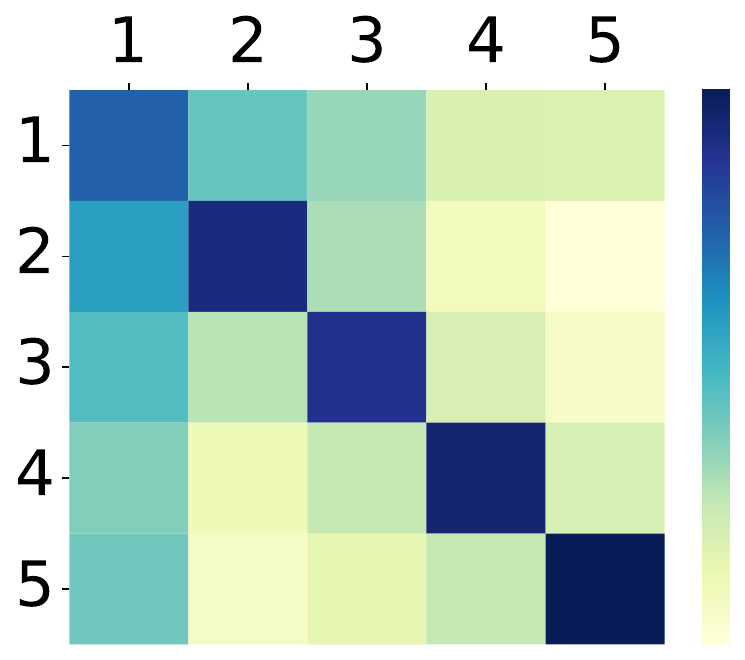}
    }
    \caption{The focus shifting phenomenon.}
    \label{fig:shifting}
\end{figure*}

To analyze the focus shifting phenomenon within the datasets, we employ BGE embedding\footnote{\scriptsize\url{https://huggingface.co/BAAI/bge-base-en-v1.5}}\cite{bge_embedding} to represent each sentence from the LLM's outputs for all questions. Subsequently, we calculate the cosine similarity between these embedded sentences to construct a similarity matrix for each question. An average of all these similarity matrices is computed, and the results are depicted in \cref{fig:shifting}.

The color intensity on the diagonal of the matrix is the strongest, which signifies a high degree of similarity between each sentence and itself. Moreover, there is a noticeable gradation in color intensity as the distance between the current and the target sentences increases, indicating a decrease in similarity. This pattern illustrates that the content focus of the sentences tends to diverge significantly as the LLM continues the generation, which is also called the focus shifting phenomenon\cite{lan2021modeling, sun2023history}.

\section{Use of Different NLI models}\label{sec:nli}
To further validate the robustness of our method, we conducted additional experiments using two distinct NLI models for inference and evaluation, respectively. 
Specifically, we employed the \texttt{t5\_11b\_trueteacher\_and\_anli}\footnote{\scriptsize\url{https://huggingface.co/google/t5_11b_trueteacher_and_anli}} as the verifier within our VTG system.
The \texttt{t5\_xxl\_true\_nli\_mixture}\footnote{\scriptsize{\url{https://huggingface.co/google/t5_xxl_true_nli_mixture}}} was used for evaluation, consistent with the evaluation methodology used in ALCE.
We randomly sampled 300 questions from ASQA and NQ datasets and conducted experiments using Vicuna-13B-v1.5-16K as the base LLM. 
From the results shown in \cref{tab:nli_results}, we can find that our method can outperform all the baselines in terms of both correctness and citation quality, validating the superiority of our method.

\begin{table*}[tb]
\centering
\resizebox{\linewidth}{!}{
\begin{tabular}{lcccccccccc}
\toprule
\textbf{Methods} & \multicolumn{5}{c}{\textbf{ASQA}} & \multicolumn{4}{c}{\textbf{NQ}} \\
\cmidrule(lr){2-6} \cmidrule(lr){7-10}
& \textbf{Correct.EM} & \textbf{Correct.D-F1} & \textbf{Citation.Rec} & \textbf{Citation.Prec} & \textbf{Citation.F1} & \textbf{Correct.EM} & \textbf{Citation.Rec} & \textbf{Citation.Prec} & \textbf{Citation.F1} \\
\midrule
\vani{}    & 30.69 & 26.95 & 70.31 & 59.51 & 64.46 & 55.00 & 70.94 & 61.20 & 65.71 \\
\summ{}       & 36.26 & 27.41 & 57.18 & 60.44 & 58.77 & 56.67 & 63.28 & 63.65 & 63.47 \\
\snippet{}    & 35.50 & 25.92 & 62.64 & 57.57 & 60.00 & 55.67 & 55.51 & 55.71 & 55.61 \\
\rerank{}     & 34.16 & 26.53 & 87.75 & 73.34 & 79.90 & 57.00 & 83.92 & 72.85 & 78.00 \\
\posthoc{} & 27.80 & 22.78 & 34.40 & 23.30 & 27.78 & 43.00 & 34.02 & 23.32 & 27.67  \\
\refinecite{} & 36.04 & 28.68 & 73.92 & 72.50 & 73.20 & 54.40 & 68.40 & 65.24 & 66.78 \\
\vericite{}   & 31.89 & 27.97 & 77.50 & 56.50 & 65.35 & 55.80 & 68.37 & 49.63 & 57.51 \\
\verirefine{} & 30.04 & 26.40 & 75.67 & 64.83 & 69.83 & 55.20 & 74.93 & 61.21 & 67.38 \\
\ours{}        & \textbf{40.28} & \textbf{29.22} & \textbf{92.97} & \textbf{88.00} & \textbf{90.42} & \textbf{59.00} & \textbf{87.24} & \textbf{80.66} & \textbf{83.82} \\
\bottomrule
\end{tabular}
}
\caption{Performance comparison across ASQA and NQ datasets}
\label{tab:nli_results}
\end{table*}

\section{Hyper-parameters}\label{sec:hyps}
The detailed hyper-parameters used in \ours{} for \texttt{Text-Davinci-003} and \texttt{Vicuna-13B-v1.5-16k} are shown in \cref{tab:hyps_davinci} and \cref{tab:hyps_vicuna}, respectively.

\section{Algorithm}
The algorithm procedural of \ours{} is shown in \cref{alg:VTG}

\section{Prompts}
The prompts used in our experiments are listed as follows.
It's worth noting that the prompts for \vani{} and \rerank{} are identical, so we only present the one for \vani{}. 
\begin{figure*}[t]
\begin{prompt}[title={Prompt for Sentence Generator}, label=prompt:search]
Instruction: Write an accurate, engaging, and concise answer for the given question using only the provided search results (some of which might be irrelevant).\\
Question: \{Question\} \\
Document: \{Document\} \\
Answer:
\end{prompt}
\end{figure*}

\begin{figure*}[t]
\begin{prompt}[title={Prompt for Citation Generator}, label=prompt:search]
Instructions: You will be provided with a sentence and several related documents. Your task is to directly append citation annotations to the sentence using these documents without changing the sentence. When citing documents, use [1][2][3]. \\
Cite at least one document and at most three documents. If multiple documents support the sentence, only cite a minimum sufficient subset of the documents. \\
Document:  \{Document\} \\
Sentence: \{Sentence\} \\
Sentence with citation:
\end{prompt}
\end{figure*}

\begin{figure*}[t]
\begin{prompt}[title={Prompt for Query Generation}, label=prompt:search]
Given the original question: \{Question\}.\\
The context is as follows: \{Context\}.\\
The claim is: \{Claim\}.\\
Please generate up to \{qg\_num\} questions that can help verify the claim with the following constraints:\\
1. You should output no more than \{qg\_num\} questions.\\
2. The generated questions should be diverse and focus on different aspects of the given claim.\\
Generated questions:
\end{prompt}
\end{figure*}

\begin{figure*}[t]
\begin{prompt}[title={Prompt for \vani{}}, label=prompt:search]
Instruction: Write a high-quality answer for the given question using only the provided search results and cite them properly using [1][2][3].\\
Question: \{Question\} \\
Document: \{Document\} \\
Answer:
\end{prompt}
\end{figure*}

\begin{figure*}[t]
\begin{prompt}[title={Prompt for \summ{}}, label=prompt:search]
\#\# Step 1: First Summarize the documents \\
Summarize the following document within 50 words with the question of interest \{Question\}\\
Return "irrelevant" if the document is ``irrelevant" to the question. Try to keep all the important dates, numbers, and names.\\
Title: \{Title\}\\
Text: \{Text\}\\
Summary: \\

\#\# Step 2: Generate the response based on the summary \\
Instruction: Write a high-quality answer for the given question using only the provided search results and cite them properly using [1][2][3].\\
Question: \{Question\} \\
Document: \{Document\} \\
Answer:
\end{prompt}
\end{figure*}

\begin{figure*}[t]
\begin{prompt}[title={Prompt for \snippet{}}, label=prompt:search]
\#\# Step 1: First extract relevant snippet from the documents \\
Given the following passage and the question \{Question\}, extract a useful span from the passage that can answer the question.\\
Resolve all the coreference issues to make the extracted span understandable and standalone. If the passage is not helpful for answering the question, return ``irrelevant". If there are multiple spans, merge them and only output one paragraph.\\
Title: \{Title\}\\
Text: \{Text\}\\
Extracted span: \\

\#\# Step 2: Generate the response based on the snippet \\
Instruction: Write a high-quality answer for the given question using only the provided search results and cite them properly using [1][2][3]. \\
Question: \{Question\} \\
Document: \{Document\} \\
Answer:
\end{prompt}
\end{figure*}

\begin{figure*}[t]
\begin{prompt}[title={Prompt for Citation Evaluation}, label=prompt:citation]
**Role: Data Annotator** \\
**Instructions:** \\
You are provided with the following materials:\\
- **Passage**: {passage} \\
- **Sentence**: {sentence} \\
**Task**: Assess whether the passage fully supports the sentence. \\
**Choices**: \\
1. **Fully Supports**: Select this option if the passage completely and clearly supports every aspect of the sentence. \\
2. **Does Not Fully Support**: Select this option if any discrepancies, omissions, or inaccuracies in the passage prevent it from fully supporting the sentence. \\
**Output**: \\
- If the passage fully supports the sentence, output "Yes." \\
- If it does not, output "No." \\
**Note**: Please refrain from adding any content not requested in the instructions.\\
\end{prompt}
\end{figure*}

\end{document}